%% file: main.tex
\documentclass[runningheads]{llncs}
\usepackage[T1]{fontenc}
\usepackage{graphicx}
\usepackage{hyperref}
\usepackage{color}

\usepackage{amsmath}
\usepackage{marvosym}
\usepackage{bm}
\usepackage{booktabs}
\usepackage{multirow}
\usepackage{makecell}
\usepackage{xcolor,soul,xspace}
\usepackage{color}
\usepackage{colortbl}
\usepackage{cite}
\usepackage{caption}
\usepackage{subcaption}

\begin{document}
\include{define}
\title{Flexible Multimodal Neuroimaging Fusion for Alzheimer's Disease Progression Prediction}

\author{
Benjamin Burns\inst{1} \and
Yuan Xue\inst{1,2} \and
Douglas W. Scharre\inst{3} \and
Xia Ning\inst{1,2,4}\Letter
}
\authorrunning{B. Burns et al.}
\titlerunning{Flexible Multimodal Neuroimaging Fusion for AD Progression Prediction}
\institute{
Department of Computer Science and Engineering, The Ohio State University, Columbus, USA \\
\and
Department of Biomedical Informatics, The Ohio State University, Columbus, USA \\
\and
Department of Neurology, The Ohio State University, Columbus, USA \\
\and
Translational Data Analytics Institute, The Ohio State University, Columbus, USA \\
\email{\{burns.1241, ning.104\}@osu.edu} \\
\email{\{yuan.xue, doug.scharre\}@osumc.edu}
}
\maketitle

\begin{abstract}
Alzheimer’s disease (AD) is a progressive neurodegenerative disease 
with high inter-patient variance in rate of cognitive decline.
AD progression prediction aims to forecast patient cognitive decline
and benefits from incorporating multiple neuroimaging modalities.
However, existing multimodal models fail to make 
accurate predictions when many modalities are missing during inference, 
as is often the case in clinical settings.
To increase multimodal model flexibility under high modality missingness, 
we introduce \method{},
a novel sparse mixture-of-experts method
that uses independent routers for each modality 
in place of the conventional, single router.
Using T1-weighted MRI, FLAIR, amyloid beta PET, and tau PET neuroimaging data
from the Alzheimer’s Disease Neuroimaging Initiative (ADNI), 
we evaluate \method{}, state-of-the-art \flexmoe{}, and unimodal neuroimaging models
on predicting two-year change in Clinical Dementia Rating-Sum of Boxes (\cdr) 
scores under varying levels of modality missingness.
\method{} outperforms the state of the art 
in most variations of modality missingness
and demonstrates more effective utility of 
experts than \flexmoe{}.

\end{abstract}
%

\section{Introduction}
%
Alzheimer’s disease (AD) is a progressive neurodegenerative disease 
with no known cure, and it is the most common cause of dementia~\cite{alz_2024_2024}. 
In the United States, one in nine adults over the age of 65 
have AD, and over 100,000 people die from AD annually~\cite{breijyeh_comprehensive_2020}. 
AD progression varies widely between patients, 
with some slowly progressing from mild cognitive impairment (MCI) 
to dementia while others demonstrate more rapid cognitive decline~\cite{soto_rapid_2008}. 
Consequently, it is essential to develop methods for 
AD progression prediction so that clinicians can 
provide their patients with more timely and individualized management.
Many deep learning models have been proposed 
to predict patient conversion from MCI to dementia~\cite{khatri_explainable_2023,hu_vgg-tswinformer_2023,hoang_vision_2023,al_olaimat_ppad_2023}, 
but their binary classifications only provide 
a broad indication of AD progression.
In contrast, models that directly forecast impairment progression 
offer a more nuanced understanding of a patient’s trajectory of decline 
\cite{wang_multimodal_2024,jung_deep_2024}.
Such models typically predict future values 
of continuous clinical metrics which quantify cognitive and functional impairment, 
like the Clinical Dementia Rating-Sum of Boxes (\cdr)~\cite{morris_clinical_1997}.
\cdr ranges from 0 (no impairment) to 18 (severe impairment),
with scores between 0.5 and 4 indicating MCI 
and scores greater than 4 indicating dementia~\cite{obryant_staging_2008}.
Most existing AD-related prediction methods are unimodal, 
utilizing only one neuroimaging modality~\cite{khatri_explainable_2023,hu_vgg-tswinformer_2023,altay_preclinical_2021,el-assy_novel_2024,esmaeilzadeh_end--end_2018}.
AD, however, is multifaceted,
requiring multiple neuroimaging modalities 
to effectively capture its distinct signatures \cite{kim_neuroimaging_2022}.
Structural MRI sequences like T1-weighted MRI (T1 MRI) 
and fluid attenuated inversion recovery (FLAIR) 
highlight reductions in brain tissue 
while amyloid beta PET (AB PET) and tau PET modalities 
detect protein agglomerations~\cite{aramadaka_neuroimaging_nodate}.
Multimodal models 
can fuse the complementary information 
from each neuroimaging modality, 
making them well-suited for AD-related predictions.
Unfortunately, traditional multimodal methods~\cite{castellano_automated_2024,zhou_effective_2019,odusami_explainable_2023}
suffer when some modalities supplied during model training 
are missing during inference. 
This is especially problematic in clinical contexts, where
missing modalities are common due to 
factors like perceived clinical necessity, limited resources, 
and cost barriers~\cite{lee_cost-effectiveness_2021,de_wilde_association_2018}. 
As such, the inflexibility of conventional multimodal models 
limits their applicability in clinical settings.

\begin{figure}[!tb]
\includegraphics[width=\textwidth]{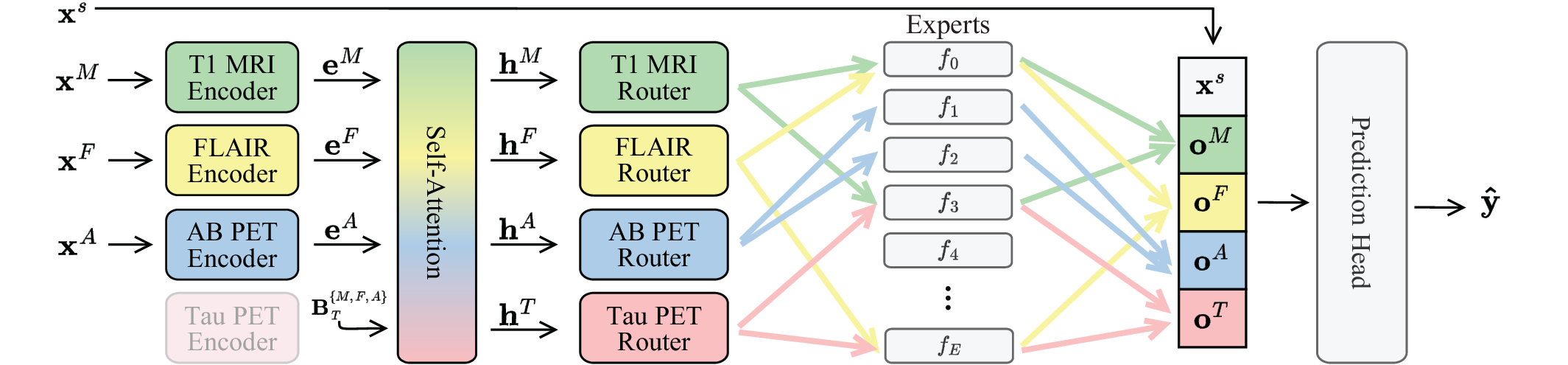}
\caption{\method{}, depicted evaluating a subject missing the tau PET modality. $\mathbf{x}^M, \mathbf{x}^F$, and $\mathbf{x}^A$ represent the 3D image data for the T1 MRI, FLAIR, and AB PET modalities, respectively. $\mathbf{e}^M, \mathbf{e}^F$, and $\mathbf{e}^A$ represent the corresponding embeddings. The tau PET embedding $\mathbf{B}_{T}^{\{M,F,A\}}$ is supplied by the missing modality bank. Modality embeddings are independently routed through the SMoE by modality-specific routers.}
\label{fig_model}
\end{figure}

Flexible multimodal modeling addresses 
the limitations of traditional multimodal models 
by increasing robustness to missing modalities. 
\flexmoe, the state of the art, utilizes a sparse mixture-of-experts (SMoE)
to achieve this.
In an SMoE, ``experts'' are simple sub-networks that are 
each trained to handle particular types of inputs.
SMoE models dynamically assign inputs to 
the most relevant subset of experts, 
enabling greater model flexibility through ensembling.
Though \flexmoe{} excels
when one or two modalities are missing,
its performance when only one modality is available remains unexplored.
To address this gap and enhance performance 
when modality availability is severely limited,
we propose \method{}, 
an improvement of \flexmoe{} that 
replaces the single router in the SMoE layer 
with multiple, modality-specific routers.
An overview of \method{} is presented in Figure~\ref{fig_model}.
In \method{}, each router independently processes inputs from 
its corresponding modality, enabling
more flexibility in routing to experts.
Using neuroimaging data from the Alzheimer's Disease Neuroimaging Initiative
(ADNI) \cite{petersen_alzheimers_2010}, we compare 
the performances of \flexmoe{}, \method{}, 
and unimodal models on predicting 
two-year change in \cdr for patients with MCI 
under various levels of modality missingness.
We demonstrate that \method{}
outperforms \flexmoe{} in most variations of modality missingness, 
achieving 8.6\%, 4.3\%, and 3.6\% improvements in RMSE 
with only FLAIR, AB PET, and tau PET, respectively.

\section{Methods}

We introduce \method{}, 
an SMoE-based multimodal fusion model 
with a novel per-modality routing mechanism
to improve robustness to missing modalities.
Built on state-of-the-art \flexmoe{}~\cite{yun_flex-moe_2024},
\method{} introduces a routing strategy that
selects a mixture of experts independently for each modality
while retaining \flexmoe{}'s core components: modality-specific 
encoders, a missing modality bank,
and an expert specialization loss.
As presented in Figure~\ref{fig_model},
model input consists of some combination of 
a patient's four neuroimaging modalities
T1 MRI, FLAIR, AB PET, and tau PET, represented as
$\mathbf{x}^{M}, \mathbf{x}^F, \mathbf{x}^A,$ and $\mathbf{x}^T$, respectively,
along with a baseline \cdr, represented as $\mathbf{x}^s$.
From these values, \method{} predicts the patient's 
two-year change in \cdr, represented as $\mathrm{\mathbf{{y}}}$.
We briefly summarize the architectural elements 
inherited from \flexmoe{} in Section~\ref{sec:flex}
and describe our novel per-modality routing method 
in Section~\ref{sec:routing}.

\subsection{\flexmoe{} Components}
\label{sec:flex}

\method builds upon state-of-the-art \flexmoe, retaining 
its modality-specific encoders, missing modality bank, 
and expert specialization loss.
For a subject with the set of available modalities $P$ 
and set of missing modalities $M$, 
modality-specific CNN encoders \cite{esmaeilzadeh_end--end_2018} 
produce embeddings for each modality $p \in P$.
Embeddings for each missing modality $m \in M$ 
are imputed as learned vectors $\mathrm{\mathbf{B}}^P_m$
conditioned on $P$.
Modality embeddings are further processed in a
self-attention layer. 
An SMoE then directs the embeddings to 
conditionally activated subsets of experts. 
SMoE outputs $\{\mathbf{o}^i\}$ for each modality $i$
are concatenated with $\mathbf{x}^s$ and
passed to an MLP prediction head to predict $\mathrm{\mathbf{{y}}}$.

\flexmoe{}'s expert specialization loss supplements 
the conventional SMoE balancing loss \cite{shazeer_outrageously_2017}.
The expert specialization loss is a cross-entropy loss that
encourages each expert to specialize in a target combination of input modalities
by routing embeddings from subjects with a given set of
available modalities to the corresponding expert.
Together, the missing modality bank and expert specialization loss
enable Flex-MoE to perform predictions on subjects with missing modalities
and specialize prediction strategies to the available set of modalities.
We refer readers to the \flexmoe{} paper \cite{yun_flex-moe_2024}
for detailed descriptions of the missing modality bank
and expert specialization loss.

\subsection{Per-Modality Routing}
\label{sec:routing}

\method improves upon \flexmoe by enhancing the SMoE routing mechanism 
for multimodal data.
Conventional SMoE models use a single router to dynamically
activate subsets of experts based on each input 
in a process known as top-$k$ routing \cite{shazeer_outrageously_2017}.
We expand top-$k$ routing to a multimodal setting 
in which each modality is assigned its own, independent router.
We term this design ``per-modality routing.''

Given an input with $M$ modalities 
where $\mathbf{h}^i$ represents 
the embedding of modality $i$ after the self-attention layer,
the SMoE output $\mathbf{o}^i$ with per-modality routing is given by 
the weighted average of $E$ expert outputs:

\begin{equation} \label{eq:smoe}
\begin{aligned}
\mathbf{o}^i &= \sum_{j=1}^E f_j(\mathbf{h}^i)  w_{ij}, \\
w_{ij} &= \topk{\router{\embed}, j}, \\
\topk{\mathbf{v}, j} &= \left\{
\begin{aligned}
  v_j  & \quad \text{if } v_j 
               \text{ is at least the }k^{\text{th}} 
               \text{ highest value in } \mathbf{v} \\
  0    & \quad \text{otherwise}
\end{aligned}
\right. \\
\router{\embed} &= \softmax{g_i(\embed)}
\end{aligned}
\end{equation}

\noindent where $g_i(\cdot)$ is the 
$i^{\text{th}}$ modality router's learned gating network, 
each $f_i$ is a learned expert network,
and $v_j$ is the $j^{\text{th}}$ element of vector $\mathbf{v}$.

When used for multimodal fusion, the single router 
in a conventional SMoE is tasked with simultaneously learning 
routing strategies across all modalities, which can be suboptimal 
due to due to differing modality-specific characteristics.
To address this, per-modality routing assigns each modality
its own router, decoupling expert selection from modality interactions
and allowing routing strategies to be learned independently for each modality.
%

\section{Materials}

\subsubsection{Dataset}

\begin{figure}[!t]
    \vspace{-5pt}
    \centering
    \begin{subfigure}[b]{0.38\textwidth}
        \includegraphics[width=\textwidth]{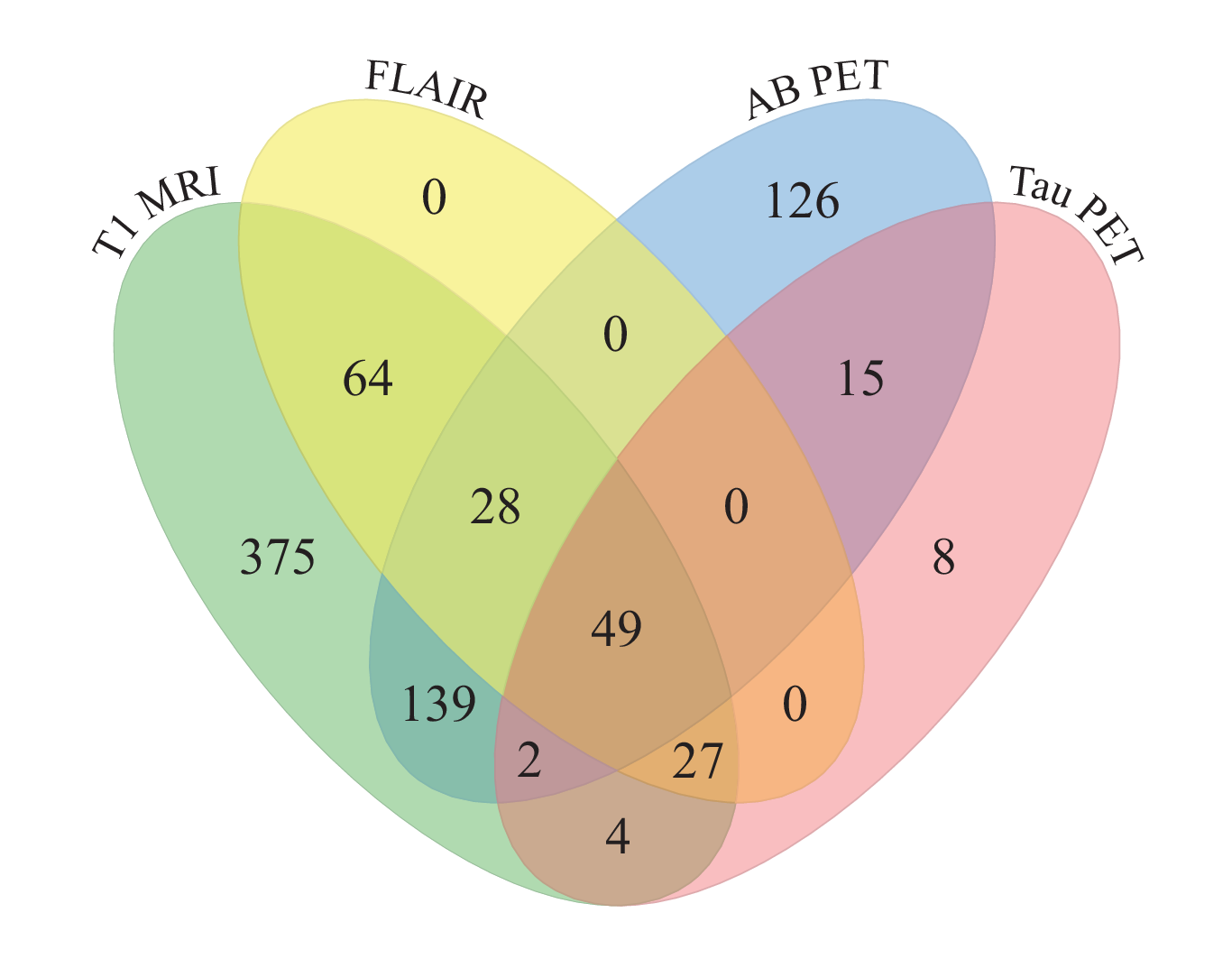}
        \caption{}
        \label{fig:vennA}
    \end{subfigure}
    \begin{subfigure}[b]{0.38\textwidth}
        \includegraphics[width=\textwidth]{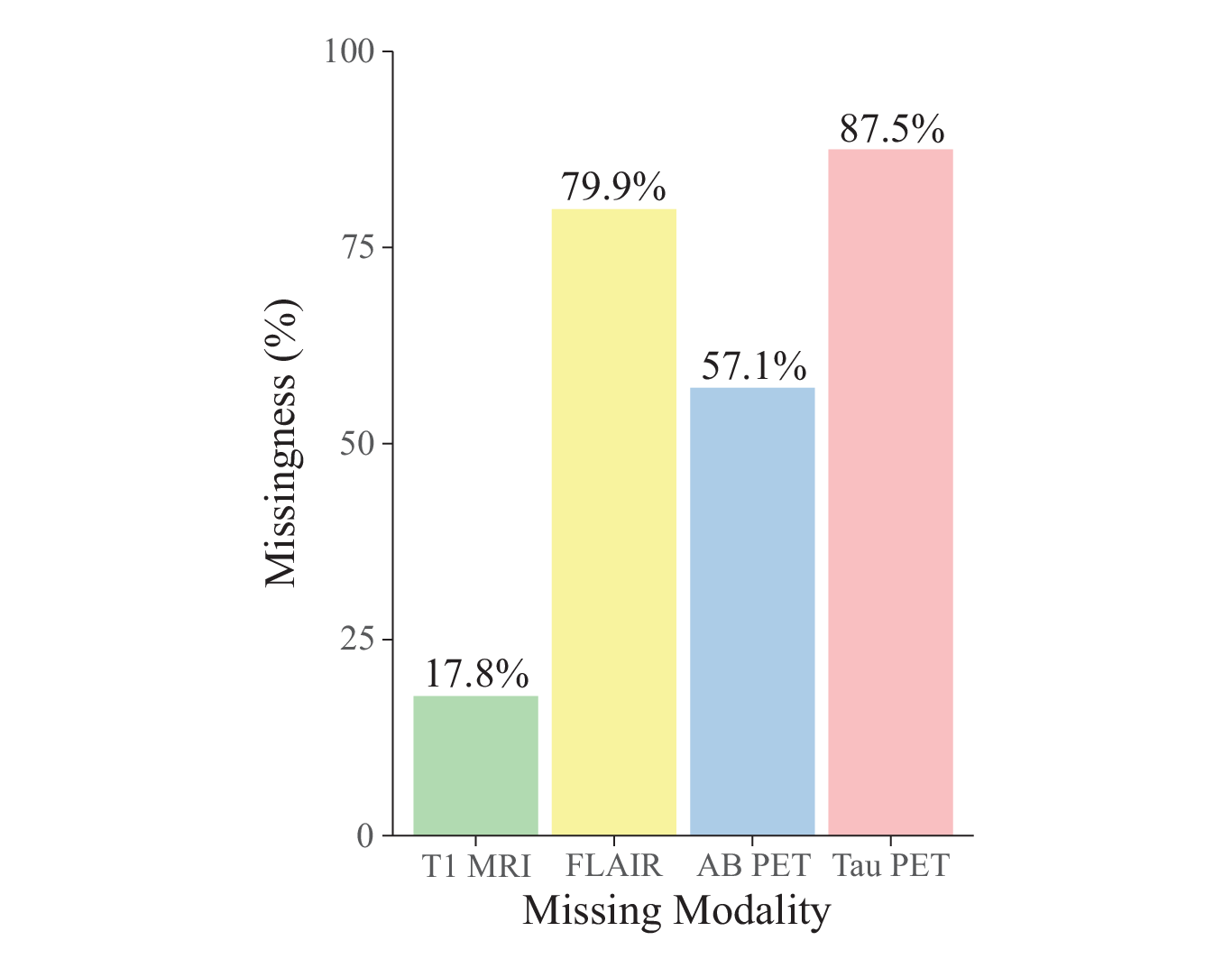}
        \caption{}
        \label{fig:vennB}
    \end{subfigure}
    \caption{(a) Venn diagram displaying the number of subjects 
    in each combination of modality availability, 
    (b) percentage of subjects missing each modality.}
    \label{fig:venn}
\end{figure}

We obtain multimodal neuroimaging and \cdr data 
from ADNI \cite{petersen_alzheimers_2010}, a longitudinal AD dataset 
tracking hundreds of participants with varying levels of cognitive impairment 
through periodic medical exams. 
From this dataset, we identify appointment pairs for each participant 
spaced two years apart.
We retain pairs that include \cdr assessments at both time points,
with at least one neuroimaging modality (T1 MRI, FLAIR, AB PET, or tau PET) 
and a \cdr above 1 at the earlier appointment.
The earlier appointment's neuroimaging modalities and \cdr 
are the subject's model input, and the change in \cdr between appointments
is the subject's target label, where each appointment pair is a subject.
In total, we obtain 837 subjects across 469 participants.
Neuroimaging modality availability varies between subjects, 
with most subjects missing at least one of the four modalities. 
Figure~\ref{fig:vennA} depicts the categorical distribution of subjects 
by available modality combination, and Figure~\ref{fig:vennB} 
displays the percentage of subjects missing each modality.
Only 5.9\% of subjects have all four modalities available, 
whereas 60.8\% of subjects have exactly one modality available.

We randomly partition the dataset into a training set (70\%), 
validation set (15\%), and testing set (15\%),
ensuring all appointment pairs from a given participant 
are grouped within a single set to prevent data leakage.
To better assess 
\method{} on inputs 
with many missing modalities, we augment the test set 
by systematically withholding neuroimaging modalities from subjects. 
For example, given a test subject that includes T1 MRI and AB PET modalities, 
we create two additional test subjects---one with only T1 MRI 
and one with only AB PET. 
In doing so, the test set is supplemented with a greater number 
of subjects with severely limited modality availability, 
better reflecting a realistic clinical setting. 

\vspace{-5pt}

\subsubsection{Neuroimaging Data Preprocessing}

Each neuroimaging modality is a collection of 3D images 
which are represented as 3D tensors.
ADNI PET images are already preprocessed so that all frames 
are averaged into a single image, and the images are 
isometrically resampled, smoothed, and standardized to 
160$\times$160$\times$96 voxels with 6mm$^3$ resolution.
We preprocess the MRI data by linearly registering each image 
to a standard template \cite{jenkinson_improved_2002}, 
removing non-brain tissues \cite{smith_fast_2002}, 
and applying N4 bias field correction \cite{tustison_n4itk_2010} 
and z-score normalization.
The resulting images are 193$\times$229$\times$193
voxels with 1mm$^3$ resolution. 
To reduce the computational cost of processing four 3D imaging modalities 
in parallel for each subject, 
we downsample all MRI and PET images by 50\% to enable 
more efficient computation with smaller modality encoders \cite{babuc_-alz_2024}.

\vspace{-5pt}

\subsubsection{Evaluation}

We assess the predictive performance of \method and \flexmoe
using root mean square error (RMSE).
For subjects with only one available modality, 
we additionally compare multimodal model performances to 
corresponding unimodal baselines, which use the same 
modality-specific encoders followed by MLP prediction heads.
We report model performance
as the average of three runs with different random parameter initializations.
Hyperparameters for each model are tuned with grid searches.
%

\section{Results and Discussion}

\subsubsection{Overall Performance}

\input{overall_performance}

We compare \method with 
state-of-the-art flexible multimodal fusion model \flexmoe
on predicting two-year change in \cdr
across varying levels of neuroimaging modality availability. 
We present our experimental results in Table~\ref{tab:main}.
\method outperforms \flexmoe{} in 11 of the 15 modality combinations, including
all combinations with only one modality available.
Specifically, \method achieves 8.6\%, 4.3\%, and 3.6\% improvements 
in RMSE over \flexmoe with only FLAIR, AB PET, and tau PET, respectively.
\method also demonstrates a 7.4\% improvement over \flexmoe 
in 5 of the 6 two-modality cases, 
with performance lagging only when 
AB PET and tau PET are available.
This combination may represent an edge case in which a
single, generalist router 
may better capture potential interactions between
these two, similar modalities.
In addition to \method's strong performance when modality availability is limited,
our method maintains competitive 2.4\% and 1.3\% average improvements with three and four
modalities, respectively.

\method's large performance gains over \flexmoe under cases of 
extreme modality missingness are likely a reflection of 
differences in the models' routing strategies.
With just a single modality on which to base predictions, 
it is crucial that the router direct embeddings 
to the most appropriate experts.
Fortunately, \method employs modality-specific routers and
is able to leverage a specialized routing strategy tailored
to the available modality.
\flexmoe, on the other hand, may suffer from poor expert selection
by its single, unspecialized router.
However, \method's performance gains become more limited 
as modality availability increases.
When multiple modalities are available,
a shared routing strategy conditioned 
on all modalities may better capture
inter-modal interactions whereas
per-modality routing may struggle to do so.

\vspace{-10pt}

\subsubsection{Performance by $\mathrm{\Delta}$\cdr}

%
\input{changecdrsb_performance}

We further analyze model performance by 
observed two-year change in \cdr, denoted as $\mathrm{\Delta}$\cdr,
in Table~\ref{tab:changecdrsb}.
We observe that \method most consistently outperforms \flexmoe
at predicting $\mathrm{\Delta}$\cdr greater than 0 and less than or equal to 1,
achieving an impressive 21.5\% average improvement in RMSE.
As such small \cdr changes typically occur early in AD progression
\cite{samtani_disease_2014},
\method{}'s strong performance for small $\mathrm{\Delta}$\cdr
demonstrates its utility in facilitating early prognosis in clinical settings.
\method also excels at predicting negative \cdr changes---improvements in
cognitive and functional impairment---achieving a 6.1\% improvement in RMSE
over \flexmoe.

\vspace{-5pt}

\subsubsection{Analysis of Expert Activation}

\begin{figure}[!t]
\centering
\includegraphics[width=.97\textwidth]{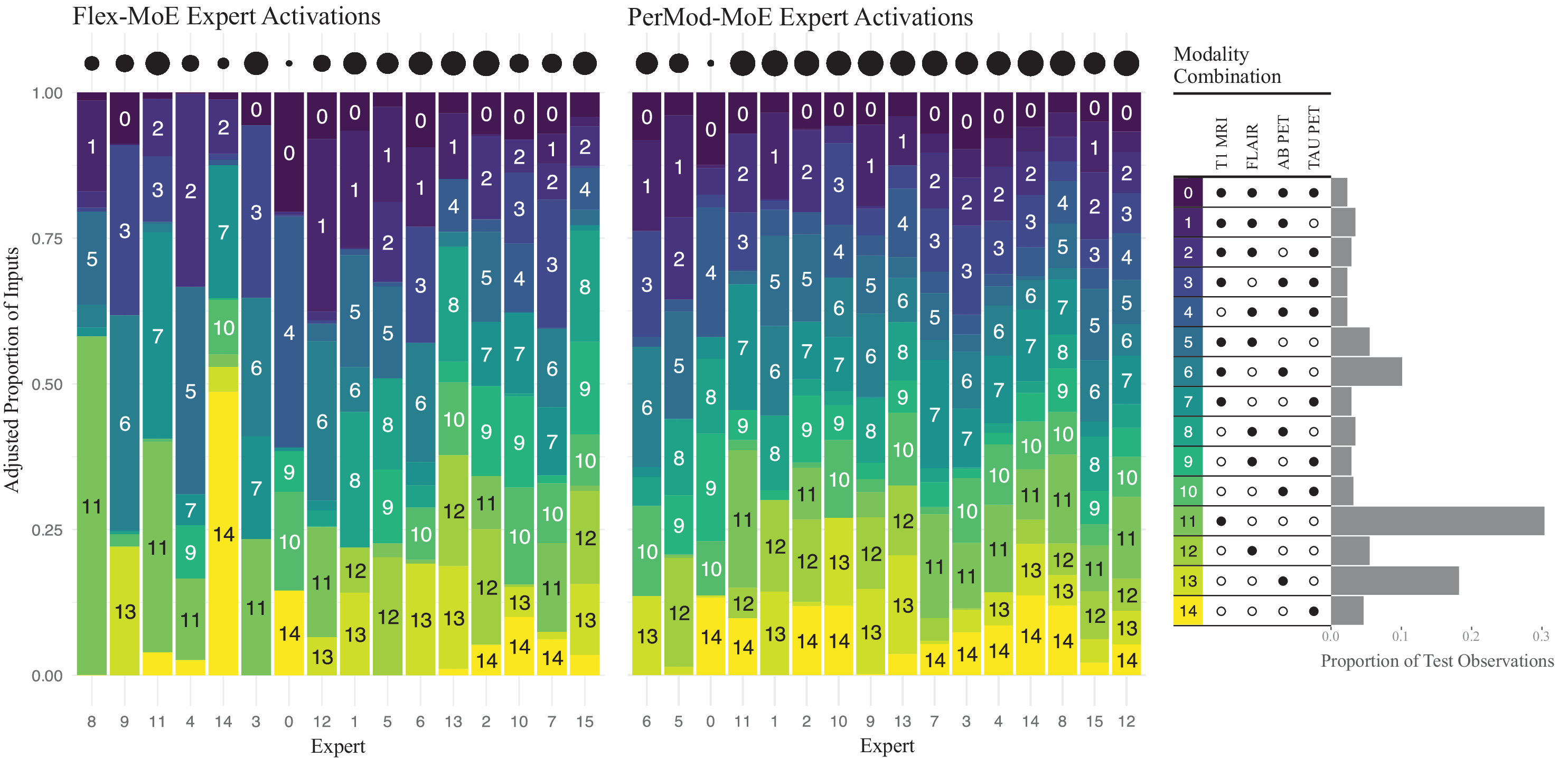}
\caption{Comparison of expert activations in \flexmoe{} and \method{}. 
Bars show the distribution of modality combinations routed to each expert; 
numbers index combinations and their intended experts (key at right). 
Expert 15 is a ``buffer'' with no target combination. 
Distributions are adjusted for variation in modality prevalence. 
Circle sizes above bars indicate how often each expert is activated, 
relative to the others, with larger sizes indicating higher activation frequencies.}
\label{fig_activation}
\vspace{-10pt}
\end{figure}

To better understand the impact of per-modality routing on expert selection,
we compare the frequencies of expert activations 
between the \method and \flexmoe in Figure~\ref{fig_activation}.
\method demonstrates a more balanced activation of experts than \flexmoe, 
with nearly all experts being selected equally.
Further, \method demonstrates more varied routing than \flexmoe, 
having more equal proportions of inputs from each modality combination 
routed to every expert. 
Still, \method displays a degree of expert specialization, 
with experts 6 and 11, for example, receiving many inputs 
from their intended modality combinations.
Based on these results, we believe that \method's performance gains 
can be attributed to its improved balance of expert activation without 
sacrificing \flexmoe's expert specialization.

\section{Conclusion}

\method{} consistently demonstrates superior performance over \flexmoe{} 
when modality availability is extremely limited 
and remains competitive with
greater modality availability. 
Given the sparsity of modality availability in realistic clinical settings, 
\method{}’s performance gains in one- and few-modality-available scenarios 
highlight its practical utility in clinical applications.
Though we evaluate \method{} using neuroimaging modalities 
to predict two-year change in \cdr prediction for patients with MCI, 
our per-modality routing design is modality-agnostic and 
can be applied in a wide variety of prediction problems.

\vspace{-5pt}

\subsubsection{Prospect of Application}
%
\method{} improves upon the state of the art in predicting AD 
progression under high modality missingness, enabling 
reliable prediction of future cognitive decline 
among people vulnerable to AD
with as few as one neuroimaging modality.
\method{}'s robustness to limited modality availability 
makes it well-suited for clinical applications, 
providing strong predictive performance without need 
for extensive and costly neuroimaging.

\begin{credits}
\subsubsection{\ackname}

This project was made possible, in part, by support from 
the National Library of Medicine grant no. R01AG082044 (X.N., D.W.S.). 
Any opinions, findings and conclusions or recommendations expressed in this paper are
those of the authors and do not necessarily reflect the views of the funding agency.

\subsubsection{\discintname}
D.W.S. is a member of the Scientific Advisory Board of BrainTest Inc.
SEZC. 
The remaining authors declare no conflicts of interest that are
relevant to the content of this article.
\end{credits}

\bibliographystyle{splncs04}
\bibliography{references}

\end{document}

%% file: define.tex
\newcommand{\method}{\mbox{PerM-MoE}\xspace}
\newcommand{\flexmoe}{\mbox{Flex-MoE}\xspace}
\newcommand{\methodgroup}{\mbox{GMod-MoE}\xspace}

\newcommand{\methodAbb}{\mbox{PerM}\xspace}
\newcommand{\flexmoeAbb}{\mbox{Flx}\xspace}
\newcommand{\uniAbb}{\mbox{Uni}\xspace}

\newcommand{\embed}{\ensuremath{\mathbf{h}^i}}
\newcommand{\topk}[1]{\ensuremath{\text{Top-}k\!\left(#1\right)\!}}
\newcommand{\softmax}[1]{\ensuremath{\text{softmax}\!\left(#1\right)\!}}
\newcommand{\router}[1]{\ensuremath{\mathcal{R}_i\!\left(#1\right)\!}}

\newcommand{\bu}{\ensuremath{\bullet}}
\newcommand{\co}{\ensuremath{\circ}}

\newcommand{\cdr}{CDR-SB\xspace}

%% file: overall_performance.tex
\newcommand{\minisize}[1]{{\fontsize{6.5}{7.5}\selectfont #1}}
\begin{table}[!tb]
    \caption{Model performances (RMSE) for two-year change in CDR-SB prediction. Variations of modality availability are indicated by the filled circles. T1 MRI, FLAIR, AB PET, and tau PET are abbreviated as M, F, A, and T, respectively. Unimodal models, \flexmoe, and \method are abbreviated as Uni, \flexmoeAbb, and \methodAbb, respectively. N is the number of test observations for each combination. The best model for each modality combination is bold and, in the unimodal cases, the second best model is underlined.}
%
    \label{tab:main}
    \centering
    \scriptsize
\setlength{\tabcolsep}{0.5pt}{
   \begin{tabular}{
   @{\hspace{0pt}}l@{\hspace{0pt}}
   @{\hspace{0pt}}l@{\hspace{0pt}}
   @{\hspace{1pt}}l@{\hspace{0pt}}
   @{\hspace{1pt}}c@{\hspace{1.75pt}}
   @{\hspace{1.75pt}}c@{\hspace{1.75pt}}
   @{\hspace{1.75pt}}c@{\hspace{1.75pt}}
   @{\hspace{1.75pt}}c@{\hspace{1.75pt}}
   @{\hspace{1.75pt}}c@{\hspace{1.75pt}}
   @{\hspace{1.75pt}}c@{\hspace{1.75pt}}
   @{\hspace{1.75pt}}c@{\hspace{1.75pt}}
   @{\hspace{1.75pt}}c@{\hspace{1.75pt}}
   @{\hspace{1.75pt}}c@{\hspace{1.75pt}}
   @{\hspace{1.75pt}}c@{\hspace{1.75pt}}
   @{\hspace{1.75pt}}c@{\hspace{1.75pt}}
   @{\hspace{1.75pt}}c@{\hspace{1.75pt}}
   @{\hspace{1.75pt}}c@{\hspace{1.75pt}}
   @{\hspace{1.75pt}}c@{\hspace{1.75pt}}
   @{\hspace{1.75pt}}c@{\hspace{0pt}}   
   }

    \toprule
    \multirow{4}{*}{}&& \textbf{M} &
        \bu & \co & \co & \co & \bu & \bu & \bu & \co & \co & \co & \bu & \bu & \bu & \co & \bu \\ && \textbf{F} &
        \co & \bu & \co & \co & \bu & \co & \co & \bu & \bu & \co & \bu & \bu & \co & \bu & \bu
        \\ && \textbf{A} &
        \co & \co & \bu & \co & \co & \bu & \co & \bu & \co & \bu & \bu & \co & \bu & \bu & \bu
        \\ && \textbf{T} &
        \co & \co & \co & \bu & \co & \co & \bu & \co & \bu & \bu & \co & \bu & \bu & \bu & \bu
        \\
    \midrule
    \uniAbb &&& 
        \makecell{\minisize{2.03} \\[-0.8ex] \tiny $\pm$0.14} &
        \makecell{\textbf{\minisize{2.15}} \\[-0.8ex] \tiny $\pm$0.06} &
        \makecell{\textbf{\minisize{2.29}} \\[-0.8ex] \tiny $\pm$0.09} &
        \makecell{\textbf{\minisize{2.11}} \\[-0.8ex] \tiny $\pm$0.19} &
        \makecell{\minisize{--}} &
        \makecell{\minisize{--}} &
        \makecell{\minisize{--}} &
        \makecell{\minisize{--}} &
        \makecell{\minisize{--}} &
        \makecell{\minisize{--}} &
        \makecell{\minisize{--}} &
        \makecell{\minisize{--}} &
        \makecell{\minisize{--}} &
        \makecell{\minisize{--}} &
        \makecell{\minisize{--}} \\
    \flexmoeAbb &&&
        \makecell{\underline{\minisize{1.95}} \\[-0.8ex] \tiny $\pm$0.09} & 
        \makecell{\minisize{2.53} \\[-0.8ex] \tiny $\pm$0.26} & 
        \makecell{\minisize{2.42} \\[-0.8ex] \tiny $\pm$0.04} & 
        \makecell{\minisize{2.56} \\[-0.8ex] \tiny $\pm$0.08} & 
        \makecell{\minisize{2.07} \\[-0.8ex] \tiny $\pm$0.15} & 
        \makecell{\minisize{2.17} \\[-0.8ex] \tiny $\pm$0.06} & 
        \makecell{\minisize{2.45} \\[-0.8ex] \tiny $\pm$0.30} & 
        \makecell{\minisize{2.06} \\[-0.8ex] \tiny $\pm$0.30} & 
        \makecell{\minisize{2.42} \\[-0.8ex] \tiny $\pm$0.07} & 
        \makecell{\textbf{\minisize{1.66}} \\[-0.8ex] \tiny $\pm$0.09} & 
        \makecell{\textbf{\minisize{1.78}} \\[-0.8ex] \tiny $\pm$0.05} & 
        \makecell{\minisize{2.39} \\[-0.8ex] \tiny $\pm$0.04} & 
        \makecell{\textbf{\minisize{2.39}} \\[-0.8ex] \tiny $\pm$0.16} & 
        \makecell{\textbf{\minisize{2.04}} \\[-0.8ex] \tiny $\pm$0.23} & 
        \makecell{\minisize{2.22} \\[-0.8ex] \tiny $\pm$0.05} \\
    \methodAbb &&& 
        \makecell{\textbf{\minisize{1.94}} \\[-0.8ex] \tiny $\pm$0.12} &
        \makecell{\underline{\minisize{2.33}} \\[-0.8ex] \tiny $\pm$0.04} &
        \makecell{\underline{\minisize{2.32}} \\[-0.8ex] \tiny $\pm$0.05} &
        \makecell{\underline{\minisize{2.47}} \\[-0.8ex] \tiny $\pm$0.14} &
        \makecell{\textbf{\minisize{1.91}} \\[-0.8ex] \tiny $\pm$0.20} &
        \makecell{\textbf{\minisize{2.01}} \\[-0.8ex] \tiny $\pm$0.06} &
        \makecell{\textbf{\minisize{2.24}} \\[-0.8ex] \tiny $\pm$0.18} &
        \makecell{\textbf{\minisize{2.03}} \\[-0.8ex] \tiny $\pm$0.12} &
        \makecell{\textbf{\minisize{2.20}} \\[-0.8ex] \tiny $\pm$0.04} &
        \makecell{\minisize{2.25} \\[-0.8ex] \tiny $\pm$0.35} &
        \makecell{\minisize{1.79} \\[-0.8ex] \tiny $\pm$0.06} &
        \makecell{\textbf{\minisize{2.06}} \\[-0.8ex] \tiny $\pm$0.17} &
        \makecell{\minisize{2.47} \\[-0.8ex] \tiny $\pm$0.31} &
        \makecell{\minisize{2.09} \\[-0.8ex] \tiny $\pm$0.11} &
        \makecell{\textbf{\minisize{2.19}} \\[-0.8ex] \tiny $\pm$0.27} \\
    \midrule
    &&\textbf{N} & 
        105 &
        19 &
        63 &
        16 &
        19 &
        35 &
        10 &
        12 &
        10 &
        11 &
        12 &
        10 &
        8 &
        8 &
        8 \\
    \bottomrule
    \end{tabular}
}
\vspace{-5pt}
\end{table}

%% file: changecdrsb_performance.tex
\begin{table}[!t]
    \vspace{-5pt}
    \centering
    \scriptsize
\setlength{\tabcolsep}{3pt}{
    \caption{Model performance comparison (RMSE) by observed two-year change in CDR-SB. Abbreviations and styling follow those in Table~\ref{tab:main}.} 
    \label{tab:changecdrsb}
   \begin{tabular}{
   @{\hspace{0pt}}c@{\hspace{2pt}}
   @{\hspace{2pt}}c@{\hspace{2pt}}
   @{\hspace{2pt}}c@{\hspace{2pt}}
   @{\hspace{2pt}}c@{\hspace{2pt}}
  @{\hspace{4pt}}c@{\hspace{4pt}}
   @{\hspace{4pt}}r@{\hspace{2pt}}
   @{\hspace{4pt}}r@{\hspace{2pt}}
   @{\hspace{2pt}}r@{\hspace{4pt}}
  @{\hspace{4pt}}c@{\hspace{4pt}}
   @{\hspace{4pt}}r@{\hspace{2pt}}
   @{\hspace{2pt}}r@{\hspace{2pt}}
   @{\hspace{2pt}}r@{\hspace{4pt}}
  @{\hspace{4pt}}c@{\hspace{4pt}}
   @{\hspace{4pt}}r@{\hspace{2pt}}
   @{\hspace{2pt}}r@{\hspace{2pt}}
   @{\hspace{2pt}}r@{\hspace{4pt}}
  @{\hspace{4pt}}c@{\hspace{4pt}}
   @{\hspace{4pt}}r@{\hspace{2pt}}
   @{\hspace{2pt}}r@{\hspace{2pt}}
   @{\hspace{2pt}}r@{\hspace{0pt}}
   } 

    \toprule

    \multicolumn{4}{c}{\textbf{\makecell{Modality\\Combo.}}}&&
    \multicolumn{3}{c}{\textbf{$\bm{\mathrm{\Delta}\text{CDR-SB}\!<\!0$}}}&&
    \multicolumn{3}{c}{\textbf{$\bm{\mathrm{\Delta}\text{CDR-SB}\!=\!0}$}}&&
    \multicolumn{3}{c}{\textbf{$\bm{0\!\!<\!\!\mathrm{\Delta}\text{CDR-SB}\!\!\leq\!\!1}$}}&&
    \multicolumn{3}{c}{\textbf{$\bm{1\!<\!\mathrm{\Delta}\text{CDR-SB}}$}}
    \\

    \cmidrule{1-4}
    \cmidrule{6-8}
    \cmidrule{10-12}
    \cmidrule{14-16}
    \cmidrule{18-20}

    M&F&A&T&&

%
%
%

   {\uniAbb}&
    {\flexmoeAbb}&
    {\methodAbb}&&

    {\uniAbb}&
    {\flexmoeAbb}&
    {\methodAbb}&&

    {\uniAbb}&
    {\flexmoeAbb}&
    {\methodAbb}&&

    {\uniAbb}&
    {\flexmoeAbb}&
    {\methodAbb}\\

    \midrule

    \bu&\co&\co&\co&&
    \textbf{1.61}&1.68&\underline{1.62}&&
    \underline{1.62}&\textbf{1.44}&1.72&&
    \textbf{1.62}&\underline{1.73}&1.94&&
    2.65&\underline{2.41}&\textbf{2.23}\\

    \co&\bu&\co&\co&&
    1.20&\textbf{0.91}&\underline{1.16}&&
    1.58&\textbf{0.81}&\underline{1.31}&&
    1.84&\textbf{1.19}&\underline{1.44}&&
    \textbf{3.13}&4.24&\underline{3.71}\\

    \co&\co&\bu&\co&&
    1.99&\textbf{1.53}&\underline{1.80}&&
    \textbf{1.65}&2.78&\underline{2.55}&&
    1.89&\underline{1.56}&\textbf{1.50}&&
    \textbf{2.96}&3.32&\underline{3.00}\\

    \co&\co&\co&\bu&&
    \textbf{2.57}&2.94&\underline{2.66}&&
    \textbf{0.59}&\underline{1.14}&1.26&&
    \textbf{1.58}&2.10&\underline{1.88}&&
    \textbf{3.07}&\underline{3.71}&3.77\\

    \cmidrule{1-20}
    
    \bu&\bu&\co&\co&&
    \multicolumn{1}{c}{--}&1.95&\textbf{1.44}&&
    \multicolumn{1}{c}{--}&1.86&\textbf{1.73}&&
    \multicolumn{1}{c}{--}&2.60&\textbf{2.41}&&
    \multicolumn{1}{c}{--}&\textbf{1.87}&2.09\\
    
    \bu&\co&\bu&\co&&
    \multicolumn{1}{c}{--}&1.33&\textbf{1.14}&&
    \multicolumn{1}{c}{--}&\textbf{0.63}&1.22&&
    \multicolumn{1}{c}{--}&3.36&\textbf{3.09}&&
    \multicolumn{1}{c}{--}&2.74&\textbf{2.48}\\

    \bu&\co&\co&\bu&&
    \multicolumn{1}{c}{--}&2.31&\textbf{1.82}&&
    \multicolumn{1}{c}{--}&\textbf{1.32}&1.56&&
    \multicolumn{1}{c}{--}&2.87&\textbf{2.23}&&
    \multicolumn{1}{c}{--}&2.85&\textbf{2.82}\\

    \co&\bu&\bu&\co&&
    \multicolumn{1}{c}{--}&\textbf{0.50}&0.73&&
    \multicolumn{1}{c}{--}&\textbf{0.45}&0.73&&
    \multicolumn{1}{c}{--}&0.70&\textbf{0.53}&&
    \multicolumn{1}{c}{--}&3.47&\textbf{3.37}\\

    \co&\bu&\co&\bu&&
    \multicolumn{1}{c}{--}&1.15&\textbf{0.83}&&
    \multicolumn{1}{c}{--}&1.76&\textbf{1.57}&&
    \multicolumn{1}{c}{--}&2.38&\textbf{1.29}&&
    \multicolumn{1}{c}{--}&\textbf{3.29}&3.45\\

    \co&\co&\bu&\bu&&
    \multicolumn{1}{c}{--}&0.69&\textbf{0.55}&&
    \multicolumn{1}{c}{--}&\textbf{1.12}&1.86&&
    \multicolumn{1}{c}{--}&\textbf{1.03}&1.78&&
    \multicolumn{1}{c}{--}&\textbf{3.27}&4.10\\

    \cmidrule{1-20}

    \bu&\bu&\bu&\co&&
    \multicolumn{1}{c}{--}&\textbf{0.40}&0.54&&
    \multicolumn{1}{c}{--}&\textbf{0.18}&0.63&&
    \multicolumn{1}{c}{--}&\textbf{0.86}&1.19&&
    \multicolumn{1}{c}{--}&2.99&\textbf{2.89}\\

    \bu&\bu&\co&\bu&&
    \multicolumn{1}{c}{--}&2.03&\textbf{1.24}&&
    \multicolumn{1}{c}{--}&1.36&\textbf{1.21}&&
    \multicolumn{1}{c}{--}&2.84&\textbf{1.49}&&
    \multicolumn{1}{c}{--}&\textbf{2.90}&3.11\\

    \bu&\co&\bu&\bu&&
    \multicolumn{1}{c}{--}&1.22&\textbf{1.06}&&
    \multicolumn{1}{c}{--}&\textbf{2.06}&2.87&&
    \multicolumn{1}{c}{--}&3.00&\textbf{2.13}&&
    \multicolumn{1}{c}{--}&\textbf{2.67}&3.17\\

    \co&\bu&\bu&\bu&&
    \multicolumn{1}{c}{--}&\textbf{0.61}&0.84&&
    \multicolumn{1}{c}{--}&\textbf{1.23}&1.93&&
    \multicolumn{1}{c}{--}&1.52&\textbf{0.99}&&
    \multicolumn{1}{c}{--}&\textbf{3.36}&3.40\\

    \cmidrule{1-20}

    \bu&\bu&\bu&\bu&&
    \multicolumn{1}{c}{--}&\textbf{0.84}&1.10&&
    \multicolumn{1}{c}{--}&\textbf{1.55}&2.32&&
    \multicolumn{1}{c}{--}&2.58&\textbf{1.63}&&
    \multicolumn{1}{c}{--}&\textbf{3.03}&3.04\\

    \bottomrule
    \end{tabular}}
    \vspace{-10pt}
\end{table}